# ANALYTICAL FORMULATIONS FOR THE LEVEL BASED WEIGHTED AVERAGE VALUE OF DISCRETE TRAPEZOIDAL FUZZY NUMBERS

Resmiye Nasiboglu[1*], Rahila Abdullayeva[2]

[1]Department of Computer Science, Dokuz Eylul University, Izmir, Turkey
[2]Department of Informatics, Sumgait State University, Sumgait, Azerbaijan

*ABSTRACT*

*In fuzzy decision-making processes based on linguistic information, operations on discrete fuzzy numbers are commonly performed. Aggregation and defuzzification operations are some of these often used operations. Many aggregation and defuzzification operators produce results independent to the decision-maker's strategy. On the other hand, the Weighted Average Based on Levels (WABL) approach can take into account the level weights and the decision maker's "optimism" strategy. This gives flexibility to the WABL operator and, through machine learning, can be trained in the direction of the decision maker's strategy, producing more satisfactory results for the decision maker. However, in order to determine the WABL value, it is necessary to calculate some integrals. In this study, the concept of WABL for discrete trapezoidal fuzzy numbers is investigated, and analytical formulas have been proven to facilitate the calculation of WABL value for these fuzzy numbers. Trapezoidal and their special form, triangular fuzzy numbers, are the most commonly used fuzzy number types in fuzzy modeling, so in this study, such numbers have been studied. Computational examples explaining the theoretical results have been performed.*

## 1. INTRODUCTION

Firstly introduced by Lotfi A. Zadeh in 1965, the fuzzy logic and fuzzy sets theory led to the integration of verbal linguistic information into mathematical models [1]. In fuzzy decision-making models based on linguistic information, usually operations on discrete fuzzy numbers are performed [2, 3].In [2], in order to merge subjective evaluations, a compensatory class of aggregation functions on the finite chain from [4] is used. Then the ranking method proposed by L. Chen and H. Lu in [5] is used to choose the best alternative, i.e., to exploit the collective linguistic preference (see [6]). This ranking method is based on the left and right dominance values of alternatives which is defined as the average difference of the left and right spreads at some discrete levels. Herein, the index of optimism is used to reflect a decision maker's degree of optimism. In our study, a more sophisticated form of this approach based on the Weighted Average Based on Levels (WABL) defuzzification operator is investigated.

Generally, defuzzification or determining the crisp representative of a fuzzy number (FN) is one of the basic operations in fuzzy inference systems, fuzzy decision-making systems and many other fuzzy logic based systems. Investigations on defuzzification methods keep their actuality

International Journal on Soft Computing (IJSC) Vol.9, No.2/3, August 2018nowadays, and various recent studies on defuzzification methods are available in the literature [7 - 10].

The well-known basic defuzzification methods are the Center of Area (COA), the Mean of Maxima (MOM), the Bisector of Area (BOA), etc. This group of methods are based on integral calculations based on the real number axis that the fuzzy number is defined. However, there are other group of methods based on integrals on [0, 1] membership degrees' axis. The most general representative of the last group of methods is the Weighted Average Based on Levels method (WABL). This method is based on the study about the mean value of the fuzzy number proposed in the pioneer study [10]. Later researches on this method have been continued in many studies [11 - 13]. More detailed investigations on the WABL approach has been handled by Nasibov with study [14] and have been continued in studies [15 - 18].

The main advantage of the WABL method is that it can be adjusted according to the decision-making strategy, or its parameters can be calculated via machine learning. In addition, the WABL parameters can be adjusted appropriately to behave as well-known methods such as COA, MOM, etc. [13, 19].In [13], one of the WABL type level based method called SLIDE is represented. The advantage of the SLIDE method is that the parameters can be adjusted to give better results in fuzzy controllers. In [13], also a machine learning approach has been given to optimally adjust the parameters of the SLIDE method. It transforms to the COA and MOM defuzzification methods as special cases.

WABL approach and its variations is used for various purposes in many other papers. In many studies, the WABL approach is handled for finding the best approximations of fuzzy numbers [20 -26]. Many other studies use the WABL approach to perform choice and ranking as well as for determining the distances between fuzzy numbers [27, 28]. In [29] an approach to obtain trapezoidal approximation of fuzzy numbers with respect to weighted distance based on WABL is proposed. In studies [30, 31] step type, and piecewise linear approximations are also investigated.

In all of the previous studies, the WABL operator is presented and investigated for fuzzy numbers with continuous universe of levels in the interval $[0, 1]$. In this study, we investigate the WABL for discrete universe of levels in the same interval:

$$\Lambda = \{\alpha_0, \alpha_1, \ldots, \alpha_t | \alpha_i \in [0, 1]; \ \alpha_0 < \alpha_1 < \cdots < \alpha_t\}. \qquad (1.1)$$

The basic forms of fuzzy numbers such as triangular and trapezoidal fuzzy numbers with discrete universe of levels and with different patterns of level weights are investigated in this study and some analytical formulas to calculate the WABL values are presented.

Rest of the paper is organized as follows. The next preliminaries section recalls the definition of the WABL operator and recalls the analytical formulas for calculation the WABL values of the continuous fuzzy numbers with various type of level weights functions. Then, in section 3, discrete leveled trapezoidal fuzzy numbers are defined and different patterns of level weight functions for discrete case are proposed. In the section 4, the levels' weights pattern functions are investigated. In the next section 5, the WABL values for discrete leveled trapezoidal FN with various levels' weights patterns are investigated and some analytical formulas are proven. Using this formulas give us a way for simple calculation of the WABL value of a fuzzy number without using more complicated integral calculations. Next, in the section 6 some computational examples calculating WABL values of discrete fuzzy numbers are illustrated. Finally, the conclusion part highlighting benefits of this study completes the paper.

## 2. PRELIMINARIES

According to the $LR$-representation, any fuzzy subset $A$ of the number axis $E$, or any fuzzy number $A$ can be defined as follows:

$$\bigcup_{\alpha \in (0,1]} (\alpha/A_\alpha) \tag{2.1}$$

where,

$$A_\alpha = [L_A(\alpha), R_A(\alpha)] = \{t \in E | L_A(\alpha) \leq t \leq R_A(\alpha)\}, \tag{2.2}$$

and $\forall \alpha \in (0,1]$, $[L_A(\alpha), R_\alpha(\alpha)]$ is a continuous closed interval. In this connection, it is assumed that $A_1 \neq \emptyset$, i.e. $A$ is a normal fuzzy number.

Let $A$ be a fuzzy number given via $LR$-representation. Density function of degrees' importance (in short – degree-importance function) we call the function $p(\alpha)$ that satisfies the following normality constraints:

$$\int_0^1 p(\alpha) d\alpha = 1, \tag{2.3}$$

$$p(\alpha) \geq 0, \ \forall \alpha \in (0,1]. \tag{2.4}$$

**Definition 2.1.** The Weighted Averaging Based on Levels (WABL) operator for a continuous fuzzy number $A$ is calculated as below:

$$WABL(A; c, p) = \int_0^1 (cR_A(\alpha) + (1-c)L_A(\alpha))p(\alpha)d\alpha, \tag{2.5}$$

where $c \in [0,1]$ is the "optimism" coefficient of the decision maker's strategy and the degree-importance function $p$ satisfies the normality constraints (2.3)-(2.4).

Based on this definition, a lot of methods can be constructed for obtaining the WABL parameters (i.e. the degree-importance function $p$ and the optimism parameter $c$). These parameters allow the method to gain flexibility. One of the methods for calculating the parameters used in WABL operator is developed with using equations system [18]. We will use the notation $WABL(A)$ instead of $WABL(A; c, p)$ for simplicity from now on.

Notice that any function $p(\alpha)$, satisfying constraints (2.3) and (2.4) could be considered as a continuous degree-importance function. The following patterns of this function is handled in studies [17, 19]:

$$p(\alpha) = (k+1)\alpha^k, \ k = 0, 1, 2, \ldots \tag{2.6}$$

It is clear that according to the parameter $k$, the degrees' importance (weights) will be constant (for $k = 0$), or be increasing linearly (for $k = 1$), quadratic (for $k = 2$), etc. w.r.t. level cuts.

In [19] it is shown that most of the well-known defuzzification operators can be simulated using the WABL operator. Simple analytical formulas to calculate WABL values of a continuous triangular and trapezoidal fuzzy numbers are formulated also in [19]. Some of these formulas are mentioned below.

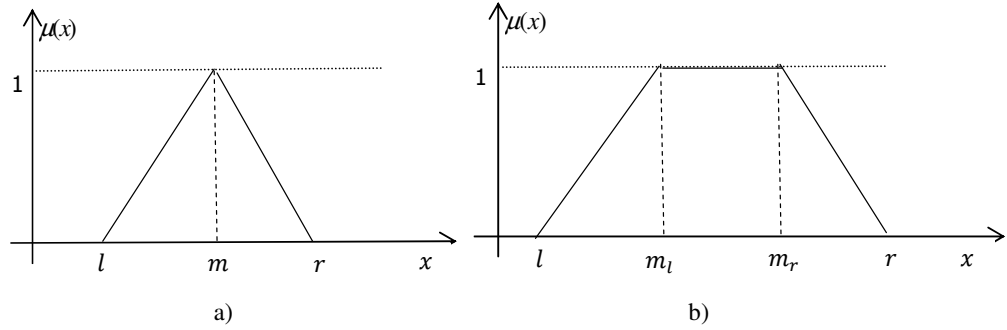

Fig. 1. a) $A = (l, m, r)$ triangular, and b) $A = (l, m_l, m_r, r)$ trapezoidal fuzzy numbers.

**Definition 2.2.** A fuzzy number with membership function in the form

$$\mu_A(x) = \begin{cases} \frac{x-l}{m-l}, & x \in [l, m), \\ \frac{r-x}{r-m}, & x \in [m, r], \\ 0, & otherwise. \end{cases} \quad (2.7)$$

is called a triangular fuzzy number $A = (l, m, r)$ (Fig. 1a).

The LR functions of the triangular fuzzy number $A = (l, m, r)$ is as follows:

$$L_A(\alpha) = l + \alpha(m - l) \text{ and } R_A(\alpha) = r - \alpha(r - m), \forall \alpha \in [0, 1] \quad (2.8)$$

**Theorem 2.1 [19].** Let a FN $A = (l, m, r)$ be a triangular fuzzy number and suppose that the distribution function of the importance of the degrees is in the form (2.6). Then the following formula for WABL is valid:

$$WABL(A) = c\left(r - \frac{k+1}{k+2}(r-m)\right) + (1-c)\left(l + \frac{k+1}{k+2}(m-l)\right), \quad (2.9)$$

where $k$ is the parameter of the degree-importance function.

**Definition 2.3.** A fuzzy number with membership function in the form

$$\mu_A(x) = \begin{cases} \frac{x-l}{m_l-l}, & x \in [l, m_l), \\ 1, & x \in [m_l, m_r) \\ \frac{r-x}{r-m_r}, & x \in [m_r, r), \\ 0, & otherwise. \end{cases} \quad (2.10)$$

is called a trapezoidal fuzzy number $A = (l, m_l, m_r, r)$ (Fig. 1b).

The *LR* functions of the trapezoidal fuzzy number $A = (l, m_l, m_r, r)$ is as follows:

$$L_A(\alpha) = l + \alpha(m_l - l) \text{ and } R_A(\alpha) = r - \alpha(r - m_r), \forall \alpha \in [0, 1] \quad (2.11)$$

**Theorem 2.2 [19].** Suppose $A = (l, m_l, m_r, r)$ is a trapezoidal fuzzy number and let the distribution function of the importance of the degrees is in the form (2.6). Then the following formula is valid for the WABL:

$$WABL(A) = c\left(r - \frac{k+1}{k+2}(r - m_r)\right) + (1 - c)\left(l + \frac{k+1}{k+2}(m_l - l)\right), \quad (2.12)$$

where $k$ is the parameter of the degree-importance function.

## 3. WABL OF A DISCRETE FUZZY NUMBER

As has been mentioned above, decision-making processes based on linguistic information, mostly performs with discrete fuzzy numbers [2, 3]. In our case, discrete fuzzy numbers with a given discrete universe $U = \{x_1, x_2, \ldots, x_n | x_i \in R, i = 1, \ldots, n\}$ and for a given discrete values of the membership degrees

$$\Lambda = \{\alpha_0, \alpha_1, \ldots, \alpha_t | \alpha_i \in [0, 1];\ \alpha_0 < \alpha_1 < \cdots < \alpha_t\} \quad (3.1)$$

is handled. Such fuzzy numbers can be represented as follows:

$$A = \bigcup_{x \in U} \mu(x)/x, \quad (3.2)$$

where $\mu(x) \in \Lambda, \forall x \in U$. This form of fuzzy number we call a discrete valued fuzzy number. In case of satisfying only the constraint (3.1), we will call the FN as discrete leveled fuzzy number.

**Definition 3.1.** Discrete triangular FN $A = (l, m, r)$ is a FN with discrete universe $U$ that

$$A = \sum_{x_i \in U} \mu_A(x_i)/x_i, \quad (3.3)$$

where

$$\mu_A(x_i) = \begin{cases} \frac{x_i - l}{m - l}, & x_i \in [l, m), \\ \frac{r - x_i}{r - m}, & x_i \in [m, r], \\ 0, & \text{otherwise.} \end{cases} \quad (3.4)$$

**Definition 3.2.** Discrete trapezoidal FN $A = (l, m_l, m_r, r)$ is a FN with discrete universe $U$ that

$$A = \sum_{x_i \in U} \mu_A(x_i)/x_i, \quad (3.5)$$

where

$$\mu_A(x_i) = \begin{cases} \frac{x_i - l}{m_l - l}, & x_i \in [l, m_l), \\ 1, & x_i \in [m_l, m_r) \\ \frac{r - x_i}{r - m_r}, & x_i \in [m_r, r), \\ 0, & \text{otherwise.} \end{cases} \quad (3.6)$$

Let $A_\alpha = \{x_i \in U | \mu(x_i) \geq \alpha\}$ be the $\alpha$ level set of the fuzzy number $A$. So it will be

$$L_A(\alpha) = min\{x_i | x_i \in A_\alpha\}, \quad (3.7)$$

$$R_A(\alpha) = max\{x_i | x_i \in A_\alpha\}. \quad (3.8)$$

Let denote

$$M_A(\alpha) = (1-c)L_A(\alpha) + cR_A(\alpha), \qquad (3.9)$$

where $c \in [0,1]$ is the "optimism" coefficient of the WABL operator, and $M_A(\alpha)$ is the mean value according to the optimism coefficient $c$ for the level $\alpha$. Then the WABL value of the fuzzy number $A$ is calculated as follows:

$$WABL(A) = \sum_{\alpha \in \Lambda} p(\alpha)(cR_A(\alpha) + (1-c)L_A(\alpha)) = \sum_{\alpha \in \Lambda} p(\alpha) M_A(\alpha) \qquad (3.10)$$

$$\sum_{\alpha \in \Lambda} p(\alpha) = 1, \qquad (3.11)$$

$$p(\alpha) \geq 0, \quad \forall \alpha \in \Lambda, \qquad (3.12)$$

where $p(\alpha), \alpha \in \Lambda$, is the degree-importance mass function.

## 4. USING OF PATTERN FUNCTIONS FOR CONSTRUCTING OF DISCRETE LEVEL WEIGHTS

We will consider the discrete FN for the case where the levels' set $\Lambda$ is a discrete set on $[0,1]$, such as $\Lambda = \{\alpha_0, \alpha_1, \ldots, \alpha_t\}$. In this case, similarly to the formula (2.6), the level weights (i.e. degree-importance) can be produced according to various patterns such as constant, linear, quadratic etc. For this purpose we can use a general pattern function as follows:

$$q(\alpha_i) \equiv q_i = i^k, \quad i = 0,1,\ldots,t. \qquad (4.1)$$

It is obvious that according to the different values of the parameter $k = 0,1,2,\ldots$, it can be produced different patterns such as constant, linear, quadratic etc. The following must be taken into account for $p(\alpha_i) \equiv p_i, \; i = 0,1,\ldots,t$,

$$p_i = \frac{q_i}{Q}, \; i = 0,1,\ldots,t, \qquad (4.2)$$

where

$$Q = \sum_{i=0}^{t} q_i . \qquad (4.3)$$

It is clear that the non-negativity and normality conditions are satisfied:

$$p_i \geq 0, \; i = 0,1,\ldots,t, \qquad (4.4)$$

$$\sum_{i=0}^{t} p_i = 1. \qquad (4.5)$$

Some special cases of the level weights are handled below.

**a. The level weights are constant.** It should be $k = 0$ in the weights' pattern function,

and

$$q_i = i^0 = 1, \; i = 0,1,\ldots,t, \qquad (4.6)$$

so

$$Q = \sum_{i=0}^{t} 1 = t+1, \qquad (4.7)$$

will be satisfied. Considering the eq. (4.2), the level weights will be in the form below

$$p_i = \frac{1}{Q} = \frac{1}{t+1}, \ i = 0,1,\dots,t. \tag{4.8}$$

**b. The weights are linearly increasing w.r.t. levels.** In this case, it should be $k = 1$ in the weights' pattern function, consequently

$$q_i = i^1 = i, \ i = 0,1,\dots,t, \tag{4.9}$$

so

$$Q = \sum_{i=0}^{t} i = \frac{t(t+1)}{2}, \tag{4.10}$$

will be satisfied. Considering eq. (4.2), the level weights will be in the form below

$$p_i = \frac{i}{Q} = \frac{2i}{t(t+1)}, \ i = 0,1,\dots,t. \tag{4.11}$$

**c. The weights are quadratic increasing w.r.t. levels.** In this case, it should be $k = 2$ in the weights' pattern function, consequently

$$q_i = i^2, \ i = 0,1,\dots,t, \tag{4.12}$$

so

$$Q = \sum_{i=0}^{t} i^2 = \frac{t(t+1)(2t+1)}{6}, \tag{4.13}$$

will be provided. Considering eq. (4.2), the level weights will be in the form below

$$p_i = \frac{i^2}{Q} = \frac{6i^2}{t(t+1)(2t+1)}, \ i = 0,1,\dots,t. \tag{4.14}$$

In the next section, some analytical formulas have been developed to calculate the WABL value for a discrete trapezoidal fuzzy number in case of equal distributed discrete levels and with different weights' pattern functions.

## 5. DETERMINING OF THE WABL VALUE FOR A DISCRETE LEVELED TRAPEZOIDAL FUZZY NUMBER IN CASE OF EQUAL DISTRIBUTED LEVELS

Let us consider the given levels are equal distributed, i.e. the levels' set is $\Lambda = \{\alpha_0, \alpha_1, \dots, \alpha_t\}$ with $\Delta \alpha = const$. So the following equalities are satisfied:

$$\Delta \alpha = \frac{1}{t} \Rightarrow \alpha_i = i\Delta\alpha, i = 0,1,\dots,t, \tag{5.1}$$

**Definition 5.1.** Let consider the trapezoidal FN $A = (l, m_l, m_r, r)$. Suppose that the level sets $A^{\alpha_i}, \ i = 0,1,\dots,t$, are constructed according to the discrete values of $\alpha_i \in [0,1], \ i = 0,1,\dots,t$. Such fuzzy numbers we call trapezoidal discrete leveled fuzzy numbers (Fig. 2).

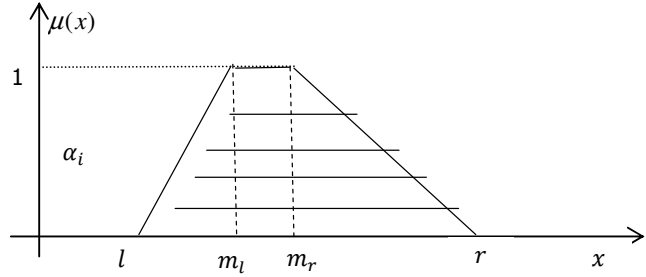

Fig. 2. The discrete leveled trapezoidal FN $A = (l, m_l, m_r, r)$ with levels $\alpha_i \in [0, 1]$, $i = 0,1,…,t$.

**Proposition 5.1.** The following equality is satisfied for a trapezoidal FN $A = (l, m_l, m_r, r)$ for any level $\alpha \in [0, 1]$:

$$M(\alpha) = M(0) + \alpha[M(1) - M(0)]. \quad (5.2)$$

**Proof:** The left and right side functions of a trapezoidal FN $A = (l, m_l, m_r, r)$ are as follows:
$$L(\alpha) = l + \alpha(m_l - l), \alpha \in [0, 1], \quad (5.3)$$

$$R(\alpha) = r - \alpha(r - m_r), \alpha \in [0, 1]. \quad (5.4)$$

So, according to eq.(3.9),

$$M(\alpha) = (1-c)L(\alpha) + cR(\alpha) = (1-c)(l + \alpha(m_l - l)) + c(r - \alpha(r - m_r)). \quad (5.5)$$

Consider that
$$M(0) = (1-c)l + cr, \quad (5.6)$$
and
$$M(1) = (1-c)m_l + cm_r, \quad (5.7)$$
we can write:
$$M(\alpha) = (1-c)l + \alpha(1-c)(m_l - l) + cr - \alpha c(r - m_r)) \quad (5.8)$$
$$= M(0) + \alpha(1-c)(m_l - l) - \alpha c(r - m_r)$$

$$= M(0) + \alpha[(1-c)(m_l - l) - c(r - m_r)]$$

$$= M(0) + \alpha[(1-c)m_l - (1-c)l - cr + cm_r]$$

$$= M(0) + \alpha[M(1) - M(0)] \quad (5.9)$$

**Proposition 5.2.** For any discrete leveled trapezoidal FN $A = (l, m_l, m_r, r)$, when $\Delta\alpha = const$, the following is valid:

$$\sum_{i=0}^{t} M(\alpha_i) = \frac{(t+1)}{2}(M(0) + M(1)) \quad (5.10)$$

**Proof:** Considering the Proposition 5.1 in case of a discrete leveled trapezoidal FN $A = (l, m_l, m_r, r)$, we can write

$$\sum_{i=0}^{t} M(\alpha_i) = \sum_{i=0}^{t}(M(0) + \alpha_i(M(1) - M(0))) =$$

$$(t+1)M(0) + (M(1) - M(0))\sum_{i=0}^{t}\alpha_i \quad (5.11)$$

Considering $\Delta\alpha = const$, we can write

$$\Delta\alpha = \frac{1}{t} \Rightarrow \alpha_i = i\Delta\alpha, i = 0,1,\dots,t, \quad (5.12)$$

So

$$\sum_{i=0}^{t}\alpha_i = \frac{1}{t}\sum_{i=0}^{t} i = \frac{1}{t}\frac{t(t+1)}{2} = \frac{t+1}{2}. \quad (5.13)$$

Considering the eq. (5.13) in (5.11), we can write:

$$\sum_{i=0}^{t} M(\alpha_i) = (t+1)M(0) + (M(1) - M(0))\frac{t+1}{2}$$
$$= \frac{(t+1)}{2}(2M(0) + M(1) - M(0))$$
$$= \frac{(t+1)}{2}(M(0) + M(1)), \quad (5.14)$$

which completes the proof.

**Proposition 5.3.** For any discrete leveled trapezoidal FN $A = (l, m_l, m_r, r)$, when $\Delta\alpha = const$, the following is valid:

$$\sum_{i=0}^{t} iM(\alpha_i) = \frac{(t+1)[3tM(0)+(2t+1)(M(1)-M(0))]}{6} \quad (5.15)$$

**Proof:**

$$\sum_{i=0}^{t} iM(\alpha_i) = \sum_{i=0}^{t} i[M(0) + \alpha_i(M(1) - M(0))]$$
$$= \sum_{i=0}^{t} iM(0) + (M(1) - M(0))\sum_{i=0}^{t} i\alpha_i \quad (5.16)$$

Considering $\alpha_i = \frac{i}{t}, i = 0,1,\dots,t,$ when $\Delta\alpha = const$, and considering the well-known following equality:

$$\sum_{i=0}^{t} i^2 = \frac{t(t+1)(2t+1)}{6}, \quad (5.17)$$

the eq. (5.16) can be continued as follows:

$$\sum_{i=0}^{t} iM(\alpha_i) = \frac{M(0)t(t+1)}{2} + (M(1) - M(0))\sum_{i=0}^{t} i\frac{i}{t}$$
$$= \frac{M(0)t(t+1)}{2} + \frac{(M(1) - M(0))}{t}\frac{t(t+1)(2t+1)}{6}$$
$$= \frac{3M(0)t(t+1) + (M(1) - M(0))(t+1)(2t+1)}{6}$$
$$= \frac{(t+1)(3tM(0)+(2t+1)(M(1)-M(0))}{6}, \quad (5.18)$$

which completes the proof.

**Theorem 5.1.** If $\Delta\alpha = const$, and the level weights are equally distributed, then the WABL of the discrete leveled trapezoidal FN $A = (l, m_l, m_r, r)$ is as follows:

$$WABL(A) = \frac{M(0)+M(1)}{2} \tag{5.19}$$

**Proof:** It is cler that

$$WABL(A) = \sum_{\alpha \in \Lambda} p(\alpha)(cR(\alpha) + (1-c)L(\alpha)) = \sum_{\alpha \in \Lambda} p(\alpha)M(\alpha). \tag{5.20}$$

We assume that $\Lambda = \{\alpha_0, \alpha_1, \dots, \alpha_t\}$, so the equation (5.16) can be written as follows:

$$WABL(A) = \sum_{i=0}^{t} p_i M(\alpha_i) \tag{5.21}$$

where $p_i \equiv p(\alpha_i), i = 0,1,\dots,t$.

In case of the equal distributed level weights, the level weights will be in the form (4.8). According to the Proposition 5.2, the following is valid

$$\sum_{i=0}^{t} M(\alpha_i) = \frac{t+1}{2}(M(0) + M(1)), \tag{5.22}$$

so we can write

$$WABL(A) = \sum_{i=0}^{t} p_i M(\alpha_i) = \frac{1}{t+1}\sum_{i=0}^{t} M(\alpha_i) = \frac{M(0)+M(1)}{2}, \tag{5.23}$$

which completes the proof.

Let us consider the linear increasing distribution of the levels' weights as in (4.9). So, the level weights must be as (4.11).

**Theorem 5.2.** If $\Delta\alpha = const$, and the level weights are linear increasing w.r.t. levels according to the pattern (4.9), then the WABL of the discrete leveled trapezoidal FN $A = (l, m_l, m_r, r)$ is as follows:

$$WABL(A) = M(0) + \frac{2t+1}{3t}\big(M(1) - M(0)\big) \tag{5.24}$$

**Proof:** Considering that

$$\Delta\alpha = const \Rightarrow \alpha_i = \frac{i}{t}, \ i = 0,1,\dots,t, \tag{5.25}$$

the level weights have the pattern (4.11), and $M(\alpha_i)$ can be calculated as in Proposition 5.1, the following equalities can be written:

$$WABL(A) = \sum_{i=0}^{t} p_i M(\alpha_i) = \frac{2}{t(t+1)}\sum_{i=0}^{t} iM(\alpha_i). \tag{5.26}$$

Considering the Proposition 5.3 that

$$\sum_{i=0}^{t} iM(\alpha_i) = \frac{(t+1)[3tM(0)+(2t+1)\big(M(1)-M(0)\big)]}{6}, \tag{5.27}$$

we can write

$$WABL(A) = \frac{2}{t(t+1)} \cdot \frac{(t+1)[3tM(0) + (2t+1)(M(1) - M(0))]}{6} =$$
$$\frac{3tM(0)+(2t+1)(M(1)-M(0))}{3t} = M(0) + \frac{2t+1}{3t}(M(1) - M(0)) \quad , \tag{5.28}$$

which completes the proof.

It is clear that when $m_l = m_r$, the trapezoidal FN becomes a triangular one. Thus, the provision of the all previous propositions and theorems are also valid for triangular fuzzy numbers.

## 6. COMPUTATIONAL EXAMPLES

The first example is about calculation of the WABL value for any discrete fuzzy number with any discrete set of levels (without the assumption that $\Delta\alpha = const$).

**Example 6.1.** Let calculate the WABL value of the discrete fuzzy number given below:

$$A = \frac{0.1}{-2} + \frac{0.4}{0} + \frac{0.7}{1} + \frac{1}{2} + \frac{0.7}{4} + \frac{0.5}{5}, \tag{6.1}$$

when the "optimism" parameter $c = 0.2$. Suppose that the levels weights are as follows:

$$p(0.1) = 0.1, \ p(0.4) = 0.3, \ p(0.5) = 0.3, \ p(0.7) = 0.2, \ p(1.0) = 0.1. \tag{6.2}$$

It is clear from the conditions of the example that the discrete universe is $U = \{-2, 0, 1, 2, 4, 5\}$ and the levels' set is $\Lambda = \{0.1, 0.4, 0.5, 0.7, 1.0\}$.

So we can calculate (let denote $M_\alpha \equiv M(\alpha)$):

$$L_{0.1} = min\{-2, 0, 1, 2, 4, 5\} = -2; \ R_{0.1} = max\{-2, 0, 1, 2, 4, 5\} = 5, \tag{6.3}$$

$$M_{0.1} = 0.8 \cdot (-2) + 0.2 \cdot 5 = -0.6; \tag{6.4}$$

$$L_{0.4} = min\{0, 1, 2, 4, 5\} = 0; \ R_{0.4} = max\{0, 1, 2, 4, 5\} = 5, \tag{6.5}$$

$$M_{0.4} = 0.8 \cdot 0 + 0.2 \cdot 5 = 1.0; \tag{6.6}$$

$$L_{0.5} = min\{1, 2, 4, 5\} = 1; \ R_{0.5} = max\{1, 2, 4, 5\} = 5, \tag{6.7}$$

$$M_{0.5} = 0.8 \cdot 1 + 0.2 \cdot 5 = 1.8; \tag{6.8}$$

$$L_{0.7} = min\{1, 2, 4\} = 1; \ R_{0.7} = max\{1, 2, 4\} = 4, \tag{6.9}$$

$$M_{0.7} = 0.8 \cdot 1 + 0.2 \cdot 4 = 1.6; \tag{6.10}$$

$$L_{1.0} = min\{2\} = 2; \ R_{1.0} = max\{2\} = 2, \tag{6.11}$$

$$M_{1.0} = 0.8 \cdot 2 + 0.2 \cdot 2 = 2.0; \tag{6.12}$$

Therefore,

$$WABL(A) = \sum_{\alpha \in \Lambda} p_\alpha M_\alpha =$$

$$0.1 \cdot (-0.6) + 0.3 \cdot 1.0 + 0.3 \cdot 1.8 + 0.2 \cdot 1.6 + 0.1 \cdot 2.0 = 1.3 \tag{6.13}$$

The next example is about calculation of the WABL for a discrete leveled trapezoidal fuzzy number with a discrete set of levels with the equally distributed level weights and with the assumption that $\Delta\alpha = const$.

**Example 6.2.** Let calculate the WABL value of the discrete leveled trapezoidal fuzzy number $A = (10, 14, 15, 23)$ (Fig. 3), and assume that the "optimism" parameter is: $c = 0.8$. Suppose that the levels are equally distributed and the levels' weights are generated according the pattern function $q_i = i^0 = 1$, $i = 0,1,...,4$, so

$$Q = \sum_{i=0}^{4} q_i = 5. \tag{6.14}$$

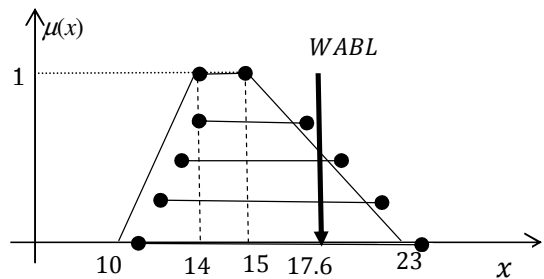

Fig.3. The discrete leveled trapezoidal FN $A = (10, 14, 15, 23)$ and its WABL value.

According to the eq. (4.5), the level weights will be in the form below

$$p_i = \frac{1}{Q} = \frac{1}{5}, \quad i = 0,1,...,4. \tag{6.15}$$

Now, we calculate the $M(0)$ and $M(1)$:

$$M(0) = (1-c)l + cr = 0.2 \cdot 10 + 0.8 \cdot 23 = 20.4, \tag{6.16}$$

$$M(1) = (1-c)m_l + cm_r = 0.2 \cdot 14 + 0.8 \cdot 15 = 14.8. \tag{6.17}$$

So we can calculate the WABL value quickly according to the theorem 5.1,

$$WABL(A) = \frac{M(0)+M(1)}{2} = \frac{20.4+14.8}{2} = 17.6 \quad . \tag{6.18}$$

Finally, the following example is about calculation of the WABL for a discrete trapezoidal fuzzy number with a discrete set of levels with the assumption that $\Delta\alpha = const$, and levels' weights are linear increasing w.r.t the levels.

**Example 6.3.** Let calculate the WABL value of the same discrete trapezoidal fuzzy number $A = (10, 14, 15, 23)$, and assume that the "optimism" parameter also is $c = 0.8$. Now suppose that the levels' weights are generated according the pattern function:

$$q_i = i^1 = i, \quad i = 0,1,...,t, \tag{6.19}$$

with $t = 4$, so

$$Q = \sum_{i=0}^{4} q_i = \sum_{i=0}^{4} i = 10. \tag{6.20}$$

According to the eq. (4.11), the level weights will be in the form below:

$$p_i = \frac{i}{Q} = \frac{i}{10}, \quad i = 0,1,\ldots,4. \tag{6.21}$$

i.e.

$$p_0 = 0, \ p_1 = \frac{1}{10}, \ p_2 = \frac{2}{10}, \ p_3 = \frac{3}{10}, \ p_4 = \frac{4}{10}. \tag{6.22}$$

It is clear that because the "optimism" parameter $c$ is the same to the example 6.2, the values of the $M(0)$ and $M(1)$ will be the same to the previous example, so $M(0) = 20.4$ and $M(1) = 14.8$. Finally, we can calculate the WABL value quickly according to the theorem 5.2,

$$WABL(A) = M(0) + \frac{(2t+1)}{3t}\big(M(1) - M(0)\big) = 20.4 + \frac{9}{12}(14.8 - 20.4) = 19.9. \tag{6.23}$$

## 7. CONCLUSION

In this study, we handle the discrete fuzzy numbers that are used in various type of fuzzy decision-making systems with linguistic information. Moreover, the trapezoidal and their special form, triangular fuzzy numbers, are the most commonly used fuzzy number types in fuzzy modeling. So in this study, such type of discrete fuzzy numbers have been considered. The WABL operator, which take into account the level weights and the decision maker's "optimism" coefficient, are defined and investigated for these numbers. Note that the flexibility of the WABL operator gives opportunity through machine learning, train its parameters according to the decision maker's strategy, producing more satisfactory results for the decision maker. In this study, simple analytical formulas have been formulated for the calculation of WABL values for discrete trapezoidal fuzzy numbers $A = (l, m_l, m_r, r)$ with constant, linear and quadratic form pattern functions of level weights. Examples, reinforcing the use of the theoretical formulas, have been demonstrated. However, since trapezoidal fuzzy number transforms to the triangular one $A = (l, m, r)$ when $m_l = m_r = m$, all the results are also valid for discrete triangular fuzzy numbers.

In our future studies, we plan to develop analytical formulas that facilitate the calculation of WABL for parametric trapezoidal discrete fuzzy numbers, which is a more general form of the trapezoidal discrete fuzzy numbers.

### ACKNOWLEDGEMENTS

This study is partially funded by Scientific Research Projects Coordination Office of DokuzEylul University under grant 2017.KB.FEN.015.